\documentclass[preprint,10pt]{elsarticle}

\usepackage{amssymb}
\usepackage{amsmath}
\usepackage{graphicx}%
\usepackage{multirow}%
\usepackage{amsmath,amssymb,amsfonts}%
\usepackage{amsthm}%
\usepackage{mathrsfs}%
\usepackage[title]{appendix}%
\usepackage{xcolor}%
\usepackage{textcomp}%
\usepackage{manyfoot}%
\usepackage{booktabs}%
\usepackage{algorithm}%
\usepackage{algorithmicx}%
\usepackage{algpseudocode}%
\usepackage{listings}%
\usepackage{times}
\usepackage{latexsym}
\usepackage{pifont}

\usepackage{amsmath,amssymb} % define this before the line numbering.
\usepackage{mathtools}
\usepackage{color}

\usepackage{algorithm}
\usepackage{algpseudocode}
\usepackage[export]{adjustbox} % Vertical alignment
\usepackage{array,multirow}
\usepackage{url}
\usepackage{tikz}
\usepackage{cleveref}

\usepackage{lineno}
\journal{Pattern Recognition}

\begin{document}

\begin{frontmatter}

%% Title, authors and addresses

%% use the tnoteref command within \title for footnotes;
%% use the tnotetext command for theassociated footnote;
%% use the fnref command within \author or \affiliation for footnotes;
%% use the fntext command for theassociated footnote;
%% use the corref command within \author for corresponding author footnotes;
%% use the cortext command for theassociated footnote;
%% use the ead command for the email address,
%% and the form \ead[url] for the home page:
%% \title{Title\tnoteref{label1}}
%% \tnotetext[label1]{}
%% \author{Name\corref{cor1}\fnref{label2}}
%% \ead{email address}
%% \ead[url]{home page}
%% \fntext[label2]{}
%% \cortext[cor1]{}
%% \affiliation{organization={},
%%             addressline={},
%%             city={},
%%             postcode={},
%%             state={},
%%             country={}}
%% \fntext[label3]{}

\title{Preserving Privacy Without Compromising Accuracy: Machine Unlearning for Handwritten Text Recognition}

%% use optional labels to link authors explicitly to addresses:
%% \author[label1,label2]{}
%% \affiliation[label1]{organization={},
%%             addressline={},
%%             city={},
%%             postcode={},
%%             state={},
%%             country={}}
%%
%% \affiliation[label2]{organization={},
%%             addressline={},
%%             city={},
%%             postcode={},
%%             state={},
%%             country={}}

% \author{} %% Author name

% %% Author affiliation
% \affiliation{organization={},%Department and Organization
%             addressline={}, 
%             city={},
%             postcode={}, 
%             state={},
%             country={}}

\author{Lei Kang, Xuanshuo Fu, Lluis Gomez, Alicia Forn\'{e}s, \\Ernest Valveny, Dimosthenis Karatzas\\
Computer Vision Center, Universitat Autònoma de Barcelona, Barcelona, Spain\\
{\tt\small \{lkang,xuanshuo,lgomez,afornes,ernest,dimos\}@cvc.uab.es}}

\begin{abstract}
{Handwritten Text Recognition (HTR) is crucial for document digitization, but handwritten data can contain user-identifiable features, like unique writing styles, posing privacy risks. Regulations such as the ``right to be forgotten'' require models to remove these sensitive traces without full retraining. We introduce a practical encoder-only transformer baseline as a robust reference for future HTR research. Building on this, we propose a two-stage unlearning framework for multihead transformer HTR models. Our method combines neural pruning with machine unlearning applied to a writer classification head, ensuring sensitive information is removed while preserving the recognition head. We also present Writer-ID Confusion (WIC), a method that forces the forget set to follow a uniform distribution over writer identities, unlearning user-specific cues while maintaining text recognition performance. We compare WIC to Random Labeling, Fisher Forgetting, Amnesiac Unlearning, and DELETE within our prune-unlearn pipeline and consistently achieve better privacy and accuracy trade-offs. This is the first systematic study of machine unlearning for HTR. Using metrics such as Accuracy, Character Error Rate (CER), Word Error Rate (WER), and Membership Inference Attacks (MIA) on the IAM and CVL datasets, we demonstrate that our method achieves state-of-the-art or superior performance for effective unlearning. These experiments show that our approach effectively safeguards privacy without compromising accuracy, opening new directions for document analysis research. Our code is publicly available at \url{https://github.com/leitro/WIC-WriterIDConfusion-MachineUnlearning}.} \begingroup
\renewcommand\thefootnote{}\footnote{Preprint version of the accepted manuscript in \textit{Pattern Recognition}, available at DOI: \url{https://doi.org/10.1016/j.patcog.2025.112411}}%
\addtocounter{footnote}{-1}%
\endgroup
\end{abstract}

% %%Graphical abstract
% \begin{graphicalabstract}
% %\includegraphics{grabs}
% \end{graphicalabstract}

% %%Research highlights
% \begin{highlights}
% \item Research highlight 1
% \item Research highlight 2
% \end{highlights}

%% Keywords
\begin{keyword}
%% keywords here, in the form: keyword \sep keyword
Handwritten Text Recognition \sep Machine Unlearning \sep Neural Pruning \sep Membership Inference Attack
%% PACS codes here, in the form: \PACS code \sep code

%% MSC codes here, in the form: \MSC code \sep code
%% or \MSC[2008] code \sep code (2000 is the default)

\end{keyword}

\end{frontmatter}

%% Add \usepackage{lineno} before \begin{document} and uncomment 
%% following line to enable line numbers
%% \linenumbers

%% main text
%%

%%%%%%%%%%%%%%%%%%%%%%%%%%%%%%%%%%%%%%%%%%%%%%%%%%%%%%%%%%%%%%%%%%%%

\begin{figure}[ht]
  \centering
  \includegraphics[width=0.95\linewidth]{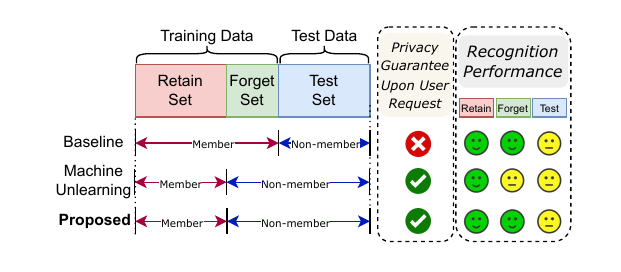}
  \vspace{-0.8cm}
  \caption{The dataset includes diverse training and test data from different writers, representing distinct domains. A baseline model is trained, and a membership inference attack reveals that the training set consists of members (green happy face), while the test set consists of non-members (yellow neutral face), highlighting a domain gap. When unlearning is requested, writer IDs are used to identify a forget set within the training data, dividing it into retain and forget sets. Existing unlearning methods aim to retain membership for the retain set and remove it from the forget and test sets, effectively erasing user data but often reducing performance. This paper introduces a method that first applies neural pruning, then performs unlearning using a writer head instead of a recognition head to forget the target data while maintaining strong performance. }
  \label{fig:first}
\end{figure}

\section{Introduction}

Handwritten Text Recognition (HTR)~\cite{plamondon2000online} has become an essential technology in the broader field of document analysis, enabling the automated extraction of textual content from handwritten sources. This capability plays a pivotal role in applications such as historical manuscript transcription~\cite{fischer2011transcription}, intelligent form processing~\cite{appalaraju2021docformer}, and digital note-taking systems~\cite{carbune2020fast}. The emergence of deep learning has significantly advanced HTR performance, allowing systems to achieve near-human accuracy in many tasks. Modern HTR models employ sophisticated neural architectures including convolutional neural networks (CNNs)~\cite{such2018fully}, recurrent neural networks (RNNs)~\cite{graves2008offline}, and Transformers~\cite{kang2022pay} to enhance both recognition accuracy and generalization across varying handwriting styles and document types. These advances have been transforming HTR from a research challenge into a deployable solution for real-world document digitization workflows.

With the increasing adoption of HTR systems, concerns surrounding privacy and data security have become more pronounced. Handwriting data inherently contains sensitive and personally identifiable information, which makes it a potential target for privacy risks when used in biometric AI applications~\cite{zhang2025privacy}. HTR models, like many other deep learning systems, often rely on large-scale datasets for training, frequently composed of user-generated content that may inadvertently include confidential or identifiable details~\cite{lukas2023analyzing}. In light of these risks, regulatory frameworks such as the European Union’s General Data Protection Regulation (GDPR)~\cite{regulation2016regulation} have been established to enforce stringent data protection requirements. GDPR obligates organizations to ensure data minimization, secure handling, and prompt deletion of user data upon request. These legal mandates introduce significant challenges for AI models that are prone to memorizing training data~\cite{carlini2021extracting, carlini2022quantifying}, thereby necessitating the development of privacy-preserving training techniques for HTR and similar systems.

A key challenge arises when attempting to remove or ``unlearn'' specific user data from a trained model without degrading its overall performance, which is the privacy-accuracy trade-off~\cite{al2019privacy}. Traditional approaches to address this issue involve retraining the model from scratch without the specified data, which is computationally expensive and impractical for large-scale systems~\cite{cao2015towards}. Recent research in machine unlearning seeks efficient methods to remove the influence of specific data points from trained models~\cite{bourtoule2021machine,kurmanji2024towards,liu2024model,kang2024machine}. However, these approaches are primarily designed for classification tasks, often leading to accuracy degradation on the forget set and posing challenges when applied to more complex tasks like HTR.

In this paper, we propose a novel approach to enable efficient machine unlearning in HTR systems without compromising accuracy. We introduce an encoder-only transformer-based model as a baseline for HTR tasks, enhanced with a handwriting style classification head. This two-task model not only serves as an indicator of how much style information (the user-identifiable component) is memorized during training but also provides a lever to do machine unlearning. By applying neural pruning on the properly trained model and subsequent unlearning techniques targeting the style classification head, we effectively remove user-identifiable information from the model. Additionally, we propose a Writer-ID Confusion (WIC) method that promotes uncertainty by enforcing a uniform distribution over writer identity predictions for the forget set, while preserving cross-entropy loss for the retain set. We further utilize membership inference attacks (MIA)~\cite{shokri2017membership} to evaluate whether user-identifiable information is eliminated as users request, demonstrating the effectiveness of our method.

Given these objectives, our study explores the following key research questions:

\textbf{[RQ1]} To what extent does the training process of an HTR model lead to the memorization of user-identifiable information?

\textbf{[RQ2]} Can neural pruning effectively remove user-identifiable information while preserving the model’s ability to recognize handwritten text?

\textbf{[RQ3]} Can employing machine unlearning in the writer classification head effectively eliminate user-identifiable information on request, without harming the recognition head’s performance?

Our main contributions are as follows:

\begin{itemize}
    \item We propose a simple encoder-only transformer-based model for the HTR task as a baseline, which can be utilized by the document analysis community for future research and development.
    \item We introduce an extra handwriting style classification head plugged to the HTR baseline model, transforming the model into a two-task architecture that performs both style classification and text recognition. This design allows us to monitor and control the memorization of user-identifiable information.
    \item We present a neural pruning method for unlearning user-identifiable information to enhance privacy. Our approach selectively removes components of the neural network by ranking the importance of the neural activations between a forget set and a retain set. By pruning the parts associated with user-specific information, we effectively eliminate personal data from the model.
    \item We propose a novel Writer-ID Confusion (WIC) method that facilitates unlearning of user-identifiable information by enforcing a uniform distribution over writer ID predictions for the forget set, while retaining standard cross-entropy training on the retain set. This introduces uncertainty in identity-specific predictions, aiding effective machine unlearning.
    \item We employ membership inference attacks to evaluate the extent of user-identifiable information memorization in the HTR model. Our extensive experiments show that after applying our unlearning method, the model effectively forgets the user-identifiable information, as indicated by the reduced success of MIA.
    \item By addressing the privacy-accuracy trade-off, our work contributes to the development of HTR systems that are both high-performing and compliant with privacy regulations. The proposed methods enable practitioners to deploy HTR models that respect user privacy without the need for costly retraining processes.
\end{itemize}

%%%%%%%%%%%%%%%%%%%%%%%%%%%%%%%%%%%%%%%%%%%%%%%%%%%%%%%%%%%%%%%%%%%%
\section{Related Work}

\textbf{Handwritten Text Recognition (HTR)} has seen remarkable improvements through the usage of deep learning techniques. Early Sequence-to-sequence approaches~\cite{dutta2018improving,kang2019convolve,chen2020multrenets,kang2021candidate} have evolved to incorporate attention mechanisms and recurrent architectures, thereby enhancing their capacity to model context. More recently, transformer-based model~\cite{kang2022pay,li2023trocr,li2025htr,fujitake2024dtrocr} have demonstrated impressive results by leveraging self-attention to capture global dependencies without the limitations of recurrent structures. Despite these advances in accuracy, however, the reliance on large volumes of user-specific handwriting data has raised significant privacy concerns.

Regulations governing \textbf{privacy and regulatory compliance in AI}, such as the EU's GDPR, enforce strict standards for data protection and the ``right to be forgotten''~\cite{voigt2017eu}. In the context of HTR systems, this mandates that models must ensure the complete removal of identifiable knowledge upon a user's request for data deletion. To the best of our knowledge, this task remains underexplored in HTR. While retraining models from scratch without the target data offers a direct solution, it is computationally intensive and impractical for large-scale applications.

\textbf{Machine Unlearning}~\cite{cao2015towards,bourtoule2021machine} has emerged as a promising approach to selectively remove information from trained models without requiring extensive retraining. Various strategies have been proposed, including gradient partitioning~\cite{yu2023unlearning}, teacher-student distillation~\cite{chundawat2023can}, influence-based removal~\cite{chen2024fast}, and pruning-based techniques~\cite{liu2024model}. However, these methods focus exclusively on classification tasks, where the objective of unlearning is often to degrade the model's performance on the target set. In contrast, HTR systems impose fundamentally different requirements, aiming to remove user-identifiable information while preserving high recognition accuracy for the target set. For instance, in the case of handwritten text images, the goal is for the model to unlearn specific writing styles or inherent lexical patterns that reveal the identity of user A, yet retain the ability to accurately recognize the textual content.

\textbf{Membership Inference Attacks (MIA)} are a pivotal tool for assessing the privacy properties of trained models. These attacks analyze model outputs to determine whether a specific data instance was part of the training set~\cite{shokri2017membership,carlini2022membership}. Although MIA research has predominantly focused on image classification and language models, it offers an essential framework for identifying privacy vulnerabilities in HTR systems.

%%%%%%%%%%%%%%%%%%%%%%%%%%%%%%%%%%%%%%%%%%%%%%%%%%%%%%%%%%%%%%%%%%%%
\section{Methodology}

\subsection{Problem Formulation}
\label{sec:problem_formulation}

The handwritten dataset $D = \{X, W, Y\}$ comprises handwritten text images $X$, associated writer identifiers $W$, and corresponding transcriptions $Y$, where each character belongs to the alphabet $A$. The alphabet $A$ consists of all English letters in both uppercase and lowercase, ranging from $A$ to $Z$ and $a$ to $z$. The dataset $D$ is partitioned into a training set and a test set, such that $D = \{D_{\text{train}}, D_{\text{test}}\}$. Furthermore, the training set is divided into a retain set and a forget set based on different writer identities, with some writers included in the retain set and others in the forget set. This division is represented as $D_{\text{train}} = \{D_{\text{retain}}, D_{\text{forget}}\}$.

\subsection{Solution Formulation}

We begin by training an HTR model $M$ using the entire training set $D_{\text{train}}$. Once the model $M$ is fully trained, it is designated as the baseline model. Next, we address the unlearning task in a scenario where a user requests the removal of user-identifiable knowledge related to a specific group of writers. This requires unlearning the handwritten data associated with the specified writers, denoted as $D_{\text{forget}}$, while maintaining high performance on the retain set $D_{\text{retain}}$. The unlearning process follows a two-stage approach: 

\textbf{Stage I: Neural Pruning}. Neural weights are selectively set to zero based on our proposed importance score to remove inherent knowledge of the forget set. 

\textbf{Stage II: Machine Unlearning}. The forget set $D_{\text{forget}}$ is further processed using machine unlearning techniques, including our proposed WIC method.

Finally, the effectiveness of the unlearning process is evaluated using the Membership Inference Attack (MIA) method $MIA$.

\subsection{Single-head Baseline Model}

To address the first research question \textbf{[RQ1]}, we introduce a baseline HTR approach that utilizes a CNN module $M_{cnn}$ to extract low-level visual features from variable-length handwritten text images, denoted as $X$. This process yields feature representations $F_{c}$, computed as $F_{c} = M_{cnn}(X)$. These extracted features are subsequently processed by a transformer-based recognizer $M_{tran}$, expressed as $F_{t} = M_{tran}(F_{c})$, which sequentially predicts the text $Y$ at the character level via a recognition head $H_r$ (implemented as a linear layer), formulated as $Y = H_r(F_{t})$. We begin by training this baseline model on the training dataset $D_{\text{train}}$. After the model has been adequately trained, we conduct a membership inference attack using the model $MIA$.

\subsection{Multi-head Baseline Model}

In the single-head framework, machine unlearning techniques such as random labeling can only be applied through the recognition head, which inevitably leads to a decline in recognition performance on the forget set. To overcome this limitation, we enhance the single-head architecture by incorporating a special $[CLS]$ token alongside the handwritten text image $X$ as input and introducing a writer classification head $H_w$ in addition to the recognition head $H_r$. This modification is formulated as follows: $F_{c}^{'} = M_{cnn}([CLS], X)$, $F_{t}^{'} = M_{tran}(F_{c}^{'})$, with the recognition head predicting the text $Y_{ocr} = H_r(F_{t}^{'})$ and the writer classification head producing the writer-specific identity $Y_{wid} = H_w(F_{t}^{'})$. By capturing user-specific attributes, such as handwriting style and inherent lexicon preferences, this design enables the model to associate such features with a unique user identity. Consequently, the writer classification head serves as an indicator of the extent to which the model retains user-specific information.

\begin{figure}
  \centering
  \includegraphics[width=0.95\linewidth]{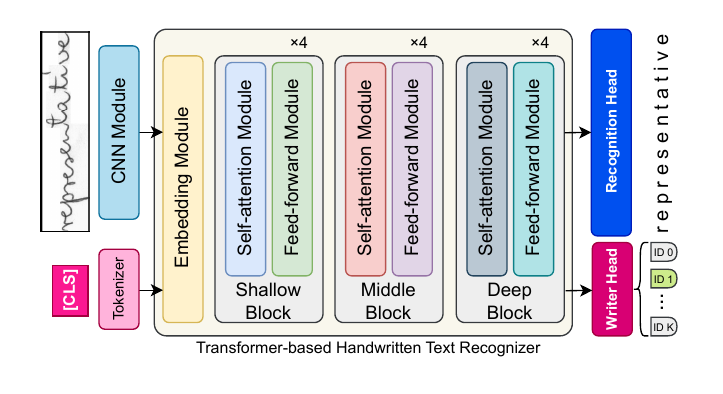}
  \caption{The architecture of the proposed multi-head transformer-based HTR method uses a $[CLS]$ special token to guide the model in projecting writer classification features through the writer classification head, while the recognition head predicts text at the character level.}
  \label{fig:arch}
\end{figure}

\subsection{Unlearning Stage I: Neural Pruning}

Based on the fully trained multi-head HTR model $M$, we first input all the handwritten text images from the retain set $D_{\text{retain}}$ into $M$ to obtain the $l$-th layer activations $S_{\text{retain}}^{l}$. Next, by feeding all the handwritten text images from the forget set $D_{\text{forget}}$ into $M$, we extract the corresponding $l$-th layer activations $S_{\text{forget}}^{l}$. The importance score for the $l$-th layer is then defined as:

\begin{equation}
    Importance\_Score = \frac{S_{\text{forget}}^{l} + \epsilon}{S_{\text{retain}}^{l} + \epsilon}
\end{equation}

where $\epsilon$ is a small constant to prevent division by zero. We then rank these importance scores for all neurons in the $l$-th layer, and set the top $K\%$ of neurons to zero. This procedure is guided by the rationale that neurons with higher importance scores are more attuned to data from the forget set, suggesting they store greater amounts of potentially sensitive information and should therefore be removed first. After the neural pruning, the update model $M^*$ is obtained.

\subsection{Unlearning Stage II: Post‑hoc Machine Unlearning}
\label{sec:stage2}

After neural pruning in Stage I, we obtain a sparsified model $M^{*}$ that has removed the most writer-sensitive neurons. In Stage II, we further eliminate residual writer-identity information while preserving the accuracy of character-level recognition. To achieve this, we investigate four state-of-the-art unlearning methods including Random Labeling, Fisher Forgetting, Amnesiac Unlearning, and DELETE, and propose a novel \emph{Writer-ID Confusion} (WIC) loss. All methods operate on mini-batches $\mathcal{B}=\mathcal{B}_{\mathrm{retain}}\cup\mathcal{B}_{\mathrm{forget}}$, with retain and forget samples defined in Sec.~\ref{sec:problem_formulation}. We denote by $\hat{\mathbf{y}}_{\mathrm{wid}}$ the logits from the writer classification head $H_{w}$, and by $\mathcal{L}_{\text{ocr}}$ the sequence-recognition loss from the baseline HTR model.

\subsubsection{Random Labeling}
\label{sec:randlabel}

We selectively corrupt the writer labels only for the forget set, transforming $D_{\text{forget}}$ into $D_{\text{forget}}^{\prime}$ through a one-to-one random permutation $perm: W_{\text{forget}} \rightarrow W \setminus W_{\text{forget}}$. The resulting training objective becomes:
\begin{equation}
\mathcal{L}_{\text{Rand}} = \text{CE}\bigl(\hat{\mathbf{y}}_{\text{wid}}, \mathbf{y}_{\text{perm}}\bigr) + \beta \mathcal{L}_{\text{ocr}},
\end{equation}
where $\mathbf{y}_{\text{perm}}$ coincides with the original labels for the retain set and the randomly permuted labels for the forget set. This approach deliberately forces misclassification of forgotten writers while minimally impacting recognition performance.

\subsubsection{Fisher Forgetting}
\label{sec:fisher}
Fisher Forgetting~\cite{golatkar2020eternal} is a method that selectively removes information about specific data points from a trained neural network. It applies a single Newton step using the Fisher Information Matrix (FIM), along with geometry-matched Gaussian noise, allowing efficient unlearning without retraining. The original formulation is as follows:

\begin{equation}
    \theta_{\text{scrub}} = \theta - (F + \mu I)^{-1} g_f + \mathcal{N}\left(0, \tau^2 (F + \mu I)^{-1}\right)
\end{equation}

In this equation, $\theta$ represents the original model parameters, $F$ is the Fisher Information Matrix estimated on the forget set, $\mu$ is a small positive damping term, $g_f$ is the gradient of the loss with respect to the forget set, and $\mathcal{N}(\cdot)$ denotes Gaussian noise with covariance scaled by the noise parameter $\tau^2$.

However, this formulation is tailored for image classification tasks such as MNIST and CIFAR10. For our sequence recognition task, we adapt the equation to a simpler form that operates on the diagonal of the Fisher Information Matrix:

\begin{equation}
    \theta_{\text{scrub}} = \theta - \alpha \frac{g_f}{F + \mu}
\end{equation}

Here, $F$ denotes the diagonal Fisher Information Matrix computed on the forget set, and $\alpha$ is the update step size (set to 0.01).

%This update is supplemented with geometry-matched Gaussian noise to facilitate unlearning while preserving general model structure.
% Fisher Forgetting~\cite{golatkar2020eternal} leverages the Fisher Information Matrix (FIM) to perform a single Newton-like parameter update. For our sequence recognition task, we employ a simplified diagonal approximation of the FIM as follows:
% \begin{equation}
% \theta_{\text{scrub}} = \theta - \alpha \frac{g_{f}}{F + \lambda},
% \label{eq:fisher_diag}
% \end{equation}
% where $g_{f}$ is the gradient computed from the forget set, $\lambda$ is a damping term, and $\alpha$ is the update step size (set to 0.01). This update is supplemented with geometry-matched Gaussian noise to facilitate unlearning while preserving general model structure.

\subsubsection{Amnesiac Unlearning}
\label{sec:amnesiac}

Amnesiac Unlearning~\cite{graves2021amnesiac} aims to reverse the training dynamics for forgotten data by iteratively applying scaled negative gradient steps:
\begin{equation}
\theta^{(t+1)} = \theta^{(t)} + \alpha \,\nabla_{\theta}\,\mathcal{L}\,\bigl(x_{f}^{t}, y_{f}^{t}\bigr),
\end{equation}
where $(x_{f}^{t},y_{f}^{t})$ is the $t$-th mini-batch drawn from $D_{\text{forget}}$, $\nabla_{\theta}$ is the gradient operator with respect to the model parameters $\theta$, and $\alpha$ is a small scaling factor (approximately $10^{-6}$). By reversing parameter updates, the method systematically erases previously learned information related to forgotten data.

\subsubsection{DELETE Distillation}
\label{sec:delete}

DELETE~\cite{zhou2025decoupled} removes sensitive classes by pairing a frozen teacher $M^T$ with a trainable student $M^{S}$. For every mini‑batch, a decoupled distillation mask $\mathbf{m}\!\in\!\mathbb{R}^{N\times K}$ sets the logits of the classes that must be forgotten to $-\infty$, leaving only the logits of the \emph{retain} classes unchanged.  
Applying \texttt{softmax} to the masked teacher outputs yields the teacher distribution
$\mathbf{p}_{r}^{T}$ that is defined solely over the retain sub‑space, ensuring no gradient ever flows through the forgotten identities.

Training minimises a single, masked knowledge‑distillation objective  
\begin{equation}
\mathcal{L}_{\text{DELETE}}
  = \lambda_{\mathrm{KD}}\;
    \mathrm{KL}\!\bigl(\mathbf{p}_{r}^{T}\,\|\,\mathbf{p}_{r}^{S}\bigr),
\end{equation}
where $\mathbf{p}_{r}^{S}$ is the student softmax restricted to the same retain classes and $\lambda_{\mathrm{KD}}$ is a balancing coefficient (set to $1$ in our experiments).  
Because the teacher is frozen and the forget‑class logits are explicitly nulled, the student converges to a state in which it no longer encodes information about the forgotten identities while preserving performance on the retain set.

\subsubsection{Fine‑Tuning}
\label{sec:finetune}

Fine-tuning (FT) is a widely used yet surprisingly strong baseline for machine unlearning. Upon the deletion request, all samples belonging to the forget classes are removed and the remaining data are used to continue training for a few additional epochs with a small learning rate. Because the procedure only sees retain data, the classifier naturally drifts away from the forgotten classes while preserving its useful knowledge.

Formally, given a mini‑batch of retain samples with one‑hot labels $\mathbf{y}_{r}$ and student logits $\mathbf{z}_{r}$ (restricted to the retain classes), FT minimises the standard cross‑entropy

\begin{equation}
\mathcal{L}_{\text{FT}}
 \;=\; 
 \mathrm{CE}\,\bigl(\hat{\mathbf{y}}_{r},\,\mathbf{y}_{r}\bigr)
\end{equation}

In our experiments, due to a budgeted number of unlearning iterations, and because the SOTA methods alone do not perform well, we first apply the SOTA methods and then use a fine-tuning strategy to achieve optimal performance.

\subsubsection{Our Proposed Writer-ID Confusion (WIC)}
\label{sec:wic}

\begin{figure}[ht]
  \centering
  \includegraphics[width=0.95\linewidth]{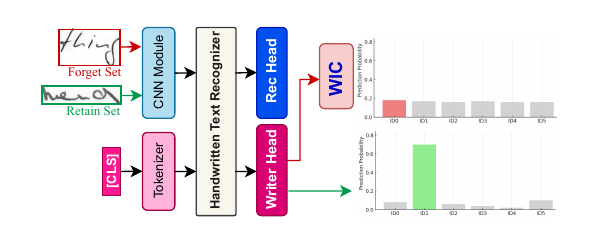}
  \caption{Pipeline of our proposed Writer-ID Confusion (WIC) method, which enforces a uniform writer ID distribution on the forget set while applying standard cross-entropy loss on the retain set.}
  \label{fig:wic}
\end{figure}

To specifically target the removal of sensitive writer identity information without compromising the model's handwritten text recognition performance, we propose the \textit{Writer-ID Confusion (WIC)} method as shown in Fig.~\ref{fig:wic}. Unlike traditional random labeling approaches, which explicitly assign incorrect writer labels, our method introduces uncertainty by enforcing a uniform distribution over writer identity predictions for samples belonging to the forget set.

Formally, the WIC loss is defined as follows:
\begin{equation}
\mathcal{L}_{\text{WIC}} = \frac{1}{|\mathcal{B}_{\text{retain}}|}\sum_{i \in \mathcal{B}_{\text{retain}}} \text{CE}(\hat{\mathbf{y}}_{\text{wid}}^{i}, y_{\text{wid}}^{i}) + \lambda \frac{1}{|\mathcal{B}_{\text{forget}}|}\sum_{j \in \mathcal{B}_{\text{forget}}}\text{KL}\left(\text{softmax}(\hat{\mathbf{y}}_{\text{wid}}^{j}) \parallel \mathbf{u}\right)
\end{equation}

Here, $\hat{\mathbf{y}}_{\text{wid}}$ represents the output logits from the writer classification head $H_w$, and $y$ denotes the true writer labels. The set $\mathcal{B}_{\text{retain}}$ consists of indices for retain set samples, while $\mathcal{B}_{\text{forget}}$ corresponds to forget set samples within a given mini-batch. The term CE represents the standard cross-entropy loss applied exclusively to the retain set to preserve the correct identification of non-sensitive writers. The KL divergence term forces the model's writer-ID predictions for forget set samples to approximate a uniform distribution, denoted by $\mathbf{u}$, effectively removing the ability to identify these writers.

The total training objective is thus formulated as:
\begin{equation}
\mathcal{L}_{\text{total}} = \mathcal{L}_{\text{WIC}} + \beta\mathcal{L}_{\text{rec}}
\end{equation}

where $\mathcal{L}\_{\text{rec}}$ is the sequence-level recognition loss, ensuring the model maintains high accuracy for handwritten text transcription. The hyperparameter $\lambda$ balances the degree of writer-ID forgetting, while setting $\beta = 1$ maintains the standard recognition performance.

\subsection{Membership Inference Attack Model}

Membership inference seeks to determine whether a specific handwritten text image was part of the training dataset for model $M$. To perform this task, we use the output logits from the recognition head as input to assess membership. These recognition output logits are expected to primarily reflect information relevant to the recognition task, rather than user-identifiable data. However, in the experimental section, we will apply MIA to evaluate this assumption. 

In this context, the retain set $D_{\text{retain}}$ is defined as the member set for membership inference, while the test set $D_{\text{test}}$ serves as the non-member set, as it was not included during training. The goal is to classify the forget set $D_{\text{forget}}$ as either belonging to the member or non-member category. Ideally, $D_{\text{forget}}$ should be classified as part of the member category for the initial training of $M$, and as part of the non-member category for the unlearned model $M'$. 

To ensure fairness in this analysis, the logits from the writer classification head are excluded. These logits are instead used to gauge whether writer information persists, as reflected in the writer classification accuracy, and to trigger the unlearning technique of random labeling.

The MIA model comprises three linear layers with a binary output, where 1 indicates a member and 0 indicates a non-member. The retain and test sets, $D_{\text{retain}}$ and $D_{\text{test}}$, are randomly partitioned into 80\% for training and 20\% for testing the MIA model. The forget set $D_{\text{forget}}$ is retained in its entirety for evaluation purposes.

%%%%%%%%%%%%%%%%%%%%%%%%%%%%%%%%%%%%%%%%%%%%%%%%%%%%%%%%%%%%%%%%%%%%
\section{Experiments}

\subsection{Implementation Details}

We implement the single- and multi-head transformer-based baseline models from scratch using PyTorch, adopting the transformer architecture from the T5 encoder. The training is conducted with a batch size of 64 and a learning rate of $2 \times 10^{-4}$, managed by a step scheduler that reduces the learning rate by 90\% every 10 epochs. The baseline models are trained for 200 epochs. In this paper, our main focus is on analyzing the relationship between privacy and accuracy. Therefore, we do not employ data augmentation or other techniques to further enhance test set performance. 

The MIA model consists of three linear layers with ReLU activation and is trained for 300 epochs. All experiments are conducted on a single NVIDIA 4090 GPU using the Adam optimization algorithm. Further details can be found in our code.

\subsection{Dataset and Metrics}

We conduct our experiments on the widely-used IAM handwritten dataset~\cite{marti2002iam}, which contains modern handwritten English texts. We utilize the RWTH partition and filter the dataset to include only upper- and lower-case letters from $a$ to $z$ and $A$ to $Z$, forming the alphabet set $A$. Our study focuses on the word level, yielding 40,977 words for training, 17,326 for validation, and 6,202 for testing. The maximum length of the output character sequence is restricted to 20. All handwritten text images are resized to a uniform height of 64 pixels while maintaining their aspect ratio, leading to variable image widths. To create mini-batches, all images are padded with blank pixels to a maximum width of 800 pixels.

The performance of the recognition task is evaluated using Character Error Rate (CER) and Word Error Rate (WER)~\cite{frinken2014continuous}, while writer classification performance is assessed using Accuracy. These metrics are defined as follows:

\begin{equation}
	   CER = \frac{S_c + I_c + D_c}{N_c}
\end{equation}

\begin{equation}
	   WER = \frac{S_w + I_w + D_w}{N_w}
\end{equation}

Here, $S$, $I$ and $D$ represent the number of substitutions, insertions, and deletions, respectively, required to transform one string into the other, either at the character or word level. $N$ denotes the total number of characters in the ground truth for CER and the total number of words in the ground truth for WER. A lower CER or WER indicates better HTR performance with fewer recognition errors.

\subsection{Single-head Baseline Analysis}
\label{sec:single}

To address the research question \textbf{[RQ1]}, we begin with initial experiments on the single-head baseline model. After completing training, the recognition performance is summarized in Tab.~\ref{tab:single_rec}. Since the forget and retain sets are part of the training set and have been observed during training, the model achieves strong performance in terms of both CER and WER. However, performance on the test set is lower due to handwriting style bias.

We conduct a membership inference evaluation, with the results summarized in Tab.~\ref{tab:single_mia}. These results clearly demonstrate that the recognition head logits can reveal user-identifiable information, as they classify samples as seen members with a 72.85\% success rate, significantly higher than the expected probability of a random guess, which is 50\%. Thus, we can address the research question \textbf{[RQ1]} by concluding that the training process of the HTR model causes it to memorize user-identifiable information.

\begin{table}[t!]
    \caption{Single-head baseline model's recognition performance.}
    %\vspace{-0.1cm}
    \label{tab:single_rec}
    \centering
    %\ 
    \small
    %\scalebox{0.99}{
    \begin{tabular}{cccccccc}
    \toprule
    \multicolumn{2}{c}{\textbf{Forget Set}} & & \multicolumn{2}{c}{\textbf{Retain Set}} & & \multicolumn{2}{c}{\textbf{Test Set}}\\
    \textbf{CER} & \textbf{WER} & & \textbf{CER} & \textbf{WER} & & \textbf{CER} & \textbf{WER}\\
    \midrule
    0.75 & 1.40 && 0.53 & 1.14 && 10.04 & 28.32\\
    \bottomrule
    \end{tabular}
    %}
    %\vspace{-0.2cm}
\end{table}

\begin{table}[t!]
    \caption{Single-head baseline model membership inference analysis.}
    %\vspace{-0.1cm}
    \label{tab:single_mia}
    \centering
    %\ 
    \small
    %\scalebox{0.99}{
    \begin{tabular}{cccccccc}
    \toprule
    \multicolumn{2}{c}{\textbf{Forget Set}} & & \multicolumn{2}{c}{\textbf{Members (Retain)}} & & \multicolumn{2}{c}{\textbf{Non-members (Test)}}\\
    \textbf{Seen} & \textbf{Unseen} && \textbf{Seen} & \textbf{Unseen} && \textbf{Seen} & \textbf{Unseen}\\
    \midrule
    72.85 & 27.15 && 80.50 & 19.50 && 47.66 & 52.34\\
    \bottomrule
    \end{tabular}
    %}
    %\vspace{-0.2cm}
\end{table}

\subsection{Multi-head Baseline Analysis}

\begin{figure}
  \centering
  \includegraphics[width=0.95\linewidth]{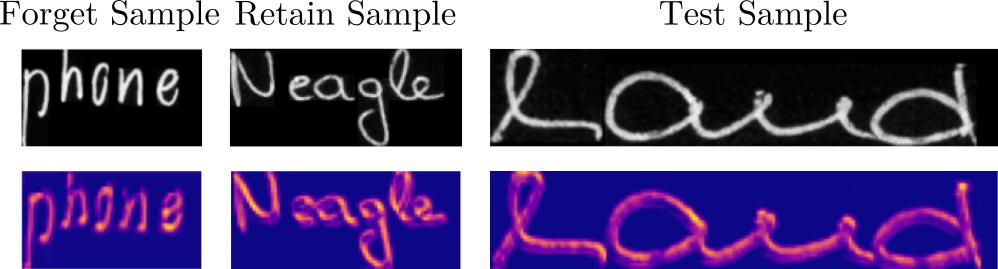}
  \caption{Handwritten text image samples and Grad-CAM visualizations are arranged for the forget, retain, and test samples from left to right, with handwritten text images and their corresponding Grad-CAM visualizations displayed from top to bottom. The groundtruth texts are ``phone'', ``Neagle'', and ``hand'' respectively.}
  \label{fig:gradcam}
\end{figure}

Based on the findings in Sec.~\ref{sec:single}, we introduce our multi-head baseline model, depicted in Fig.~\ref{fig:arch}, which incorporates a writer classification head. Across all experiments, we hypothesize that the convolutional features generated by the CNN module primarily capture low-level visual features from handwritten text images, without embedding high-level semantic information such as handwriting styles or language patterns. To support this, we employ Grad-CAM~\cite{selvaraju2017grad} to visualize the CNN features extracted by the module, as illustrated in Fig.~\ref{fig:gradcam}. From the figure, it is evident that the Grad-CAM visualizations are consistent across all three sets, proving our hypothesis that the CNN module extracts only low-level visual features from the handwritten text images, without capturing higher-level semantic information.

\subsection{Neural Pruning Experiments}

In Stage I of neural pruning, we conduct comprehensive experiments on the multi-head transformer model $M$. The embedding module combines the $[CLS]$ token with the handwritten visual feature sequence (extracted from the CNN module) to create a unified feature sequence. This sequence is then processed through 12 transformer blocks, each consisting of a self-attention module and a feed-forward module. The architecture concludes with two projection layers: one for the writer classification head and the other for the recognition head. These experiments aim to analyze the impact of neural pruning on each module, ultimately yielding a well-pruned model $M^*$ for Stage II.

\subsubsection{Full Module Pruning}

We perform neural pruning across full modules, as shown in Tab.~\ref{tab:prune_all}, applying different pruning rates to the embedding, self-attention, and feed-forward modules. The results indicate that pruning removes more information from the forget set compared to the retain set, as evidenced by a greater drop in writer classification accuracy for the forget set as pruning rates increase. In contrast, recognition performance for CER and WER shows a similar scale of decline across both sets as pruning rates increase. This suggests the need for further analysis of how each module individually impacts both writer classification accuracy and recognition performance.

\begin{table}[t!]
    \caption{Experiments with varying pruning percentages across embedding, self-attention, and feed-forward modules.}
    %\vspace{-0.1cm}
    \label{tab:prune_all}
    \centering
    %\ 
    \small
    \scalebox{0.8}{
    \begin{tabular}{cccccccccccc}
    \toprule
    \multirow{2}{*}{\textbf{Pruning Rate}} & \multirow{2}{*}{\textbf{Sparsity}} & \multicolumn{3}{c}{\textbf{Forget Set}} & & \multicolumn{3}{c}{\textbf{Retain Set}} & & \multicolumn{2}{c}{\textbf{Test Set}}\\
    & & \textbf{ACC} & \textbf{CER} & \textbf{WER} & & \textbf{ACC} & \textbf{CER} & \textbf{WER} & & \textbf{CER} & \textbf{WER}\\
    \midrule
    Original & 0\% & 100.00 & 1.80 & 1.89 && 100.00 & 1.29 & 1.75 && 13.23 & 34.90 \\
    \midrule
    5\% & 3.85\% & 99.85 & 2.47 & 3.04 && 99.86 & 1.95 & 3.13 && 15.96 & 39.70\\
    10\% & 7.70\% & 97.01 & 4.27 & 11.51 && 97.79 & 4.74 & 13.15 && 20.28 & 46.73\\
    15\% & 11.61\% & 86.55 & 12.88 & 36.27 && 91.48 & 13.67 & 37.89 && 29.02 & 59.52\\
    20\% & 15.47\% & 57.95 & 33.15 & 64.77 && 70.92 & 36.23 & 68.17 && 47.93 & 76.98\\
    25\% & 19.39\% & 33.68 & 58.38 & 84.26 && 45.04 & 60.52 & 85.70 && 68.90 & 89.39\\
    30\% & 23.24\% & 12.06 & 89.56 & 94.77 && 25.45 & 91.69 & 95.65 && 96.46 & 96.51\\
    35\% & 27.09\% & 4.98 & 97.49 & 98.26 && 16.35 & 99.86 & 98.24 && 101.89 & 98.53\\
    40\% & 31.00\% & 0.80 & 124.48 & 99.45 && 8.07 & 129.34 & 99.67 && 127.46 & 99.59\\
    \bottomrule
    \end{tabular}
    }
    %\vspace{-0.2cm}
\end{table}

\subsubsection{Embedding Module Pruning}

We apply pruning exclusively to the embedding module, as detailed in Tab.~\ref{tab:prune_embed}. The results show that as the pruning rate increases, the writer classification accuracy decreases more significantly for the forget set than for the retain set. Similarly, recognition performance, measured by CER and WER, follows a similar pattern, with slightly greater declines observed for the forget set. This indicates that pruning the embedding module results in the decline of both user-identifiable information and recognition information for both the forget and retain sets, with the forget set experiencing greater information decline.

As pruning within the embedding module increases, the decline in user-identifiable information is more pronounced in the forget set compared to the retain set. Therefore, we select a pruning rate of 40\% for the embedding module to balance the removal of information with maintaining good recognition performance.

\begin{table}[t!]
    \caption{Experiments with varying pruning percentages applied to the embedding module.}
    %\vspace{-0.1cm}
    \label{tab:prune_embed}
    \centering
    %\ 
    \small
    \scalebox{0.8}{
    \begin{tabular}{cccccccccccc}
    \toprule
    \multirow{2}{*}{\textbf{Pruning Rate}} & \multirow{2}{*}{\textbf{Sparsity}} & \multicolumn{3}{c}{\textbf{Forget Set}} & & \multicolumn{3}{c}{\textbf{Retain Set}} & & \multicolumn{2}{c}{\textbf{Test Set}}\\
    & & \textbf{ACC} & \textbf{CER} & \textbf{WER} & & \textbf{ACC} & \textbf{CER} & \textbf{WER} & & \textbf{CER} & \textbf{WER}\\
    \midrule
    Original & 0\% & 100.00 & 1.80 & 1.89 && 100.00 & 1.29 & 1.75 && 13.23 & 34.90 \\
    \midrule
    10\% & 0.14\% & 100.00 & 1.89 & 2.04 && 100.00 & 1.42 & 1.92 && 13.54 & 35.21\\
    20\% & 0.28\% & 99.95 & 2.26 & 2.89 && 99.94 & 1.67 & 2.45 && 14.17 & 36.58\\
    30\% & 0.42\% & 98.01 & 3.85 & 8.47 && 98.76 & 2.95 & 6.31 && 16.45 & 40.34 \\
    40\% & 0.56\% & 82.26 & 9.29 & 23.72 && 90.37 & 6.95 & 19.48 && 20.61 & 46.94\\
    50\% & 0.71\% & 53.36 & 17.02 & 41.01 && 73.10 & 14.13 & 36.00 && 26.68 & 55.24\\
    60\% & 0.84\% & 21.08 & 36.63 & 65.87 && 39.87 & 32.79 & 62.09 && 41.43 & 70.55\\
    70\% & 0.99\% & 5.08 & 71.18 & 88.84 && 15.78 & 67.52 & 86.51 && 70.23 & 88.02\\
    80\% & 1.13\% & 0.75 & 127.89 & 98.46 && 6.66 & 124.88 & 98.16 && 121.82 & 98.21\\
    \bottomrule
    \end{tabular}
    }
    %\vspace{-0.2cm}
\end{table}

\subsubsection{Self-attention Module Pruning}

We apply pruning exclusively to all the self-attention modules in model $M$, as shown in Tab.~\ref{tab:prune_selfattn}. The results show that as the pruning rate increases, the writer classification accuracy decreases more significantly for the forget set than for the retain set. This suggests that self-attention features in the forget set preserve more user-specific information than those in the retain set. Consequently, pruning causes a greater removal of user-specific information in the forget set. 

Nevertheless, the recognition performance, measured in terms of CER and WER, declines at a similar rate for both sets. Thus, the degradation of recognition-related information follows a similar tendency across both sets as the pruning rate increases.

\begin{table}[t!]
    \caption{Experiments with varying pruning percentages applied to the self-attention module.}
    %\vspace{-0.1cm}
    \label{tab:prune_selfattn}
    \centering
    %\ 
    \small
    \scalebox{0.8}{
    \begin{tabular}{cccccccccccc}
    \toprule
    \multirow{2}{*}{\textbf{Pruning Rate}} & \multirow{2}{*}{\textbf{Sparsity}} & \multicolumn{3}{c}{\textbf{Forget Set}} & & \multicolumn{3}{c}{\textbf{Retain Set}} & & \multicolumn{2}{c}{\textbf{Test Set}}\\
    & & \textbf{ACC} & \textbf{CER} & \textbf{WER} & & \textbf{ACC} & \textbf{CER} & \textbf{WER} & & \textbf{CER} & \textbf{WER}\\
    \midrule
    Original & 0\% & 100.00 & 1.80 & 1.89 && 100.00 & 1.29 & 1.75 && 13.23 & 34.90 \\
    \midrule
    10\% & 2.51\% & 99.10 & 2.61 & 5.83 && 99.71 & 2.71 & 5.97 && 18.12 & 43.49\\
    20\% & 5.06\% & 82.06 & 16.96 & 46.49 && 92.42 & 17.53 & 46.11 && 33.87 & 66.02\\
    30\% & 7.60\% & 42.80 & 54.90 & 86.35 && 67.31 & 55.04 & 86.72 && 64.61 & 90.21\\
    40\% & 10.15\% & 16.54 & 86.59 & 95.32 && 39.60 & 87.89 & 95.79 && 90.98 & 96.79\\
    \bottomrule
    \end{tabular}
    }
    %\vspace{-0.2cm}
\end{table}

\subsubsection{Fine-grained Pruning on self-attention Modules}

To further investigate the effect of self-attention module pruning on performance, we conduct a fine-grained pruning experiment by dividing the 12 transformer blocks into three groups: shallow blocks (0-3), middle blocks (4-7), and deep blocks (8-11). Pruning is applied to the self-attention modules within these groups, as detailed in Tab.~\ref{tab:prune_smd}. The results reveal that shallow blocks contain more user-specific and recognition-related information, as both writer classification accuracy and recognition performance (CER and WER) decrease significantly with increasing pruning rates. 

In contrast, middle blocks maintain high performance for both writer classification accuracy and recognition metrics (CER and WER) even with 40\% pruning, indicating they are less sensitive to pruning. For deep blocks, writer classification accuracy remains relatively stable from 0\% to 40\% pruning, suggesting they contain less user-specific information. However, recognition performance declines as the pruning rate increases, indicating these blocks carry more recognition-related information.

When comparing the behavior of the forget and retain sets, the trends are similar, suggesting that pruning does not remove more information from the forget set compared to the retain set. However, in the shallow blocks, the writer classification accuracy declines more significantly for the forget set than for the retain set, indicating that shallow blocks contain more user-specific information that is particularly sensitive to the forget set.

When pruning the shallow blocks, the decline of user-identifiable information is similar for both the forget and retain sets. However, increasing the pruning rate leads to a greater reduction in recognition-related information. Therefore, we choose a 20\% pruning rate for the shallow blocks. In contrast, for the deep blocks, increasing the pruning rate does not significantly affect user-identifiable information, but it does lower recognition performance. Hence, we also set a 20\% pruning rate for the deep blocks. For the middle blocks, since neither writer classification nor recognition performance significantly declines with increased pruning, we can afford to remove more knowledge. Thus, we select a 40\% pruning rate for the middle blocks.

\begin{table}[t!]
    \caption{Experiments with varying pruning percentages applied to the shallow (blocks 0-3), middle (blocks 4-7), and deep (blocks 8-11) self-attention modules.}
    %\vspace{-0.1cm}
    \label{tab:prune_smd}
    \centering
    %\ 
    \small
    \scalebox{0.8}{
    \begin{tabular}{ccccccccccccc}
    \toprule
    \multirow{2}{*}{\textbf{Layer}} & \multirow{2}{*}{\textbf{PRate}} & \multirow{2}{*}{\textbf{Sprs.}} & \multicolumn{3}{c}{\textbf{Forget Set}} & & \multicolumn{3}{c}{\textbf{Retain Set}} & & \multicolumn{2}{c}{\textbf{Test Set}}\\
    & & & \textbf{ACC} & \textbf{CER} & \textbf{WER} & & \textbf{ACC} & \textbf{CER} & \textbf{WER} & & \textbf{CER} & \textbf{WER}\\
    \midrule
    & Orig. & 0\% & 100.00 & 1.80 & 1.89 && 100.00 & 1.29 & 1.75 && 13.23 & 34.90 \\
    % & Retr. & 0\% & 0.00 & 14.09 & 37.22 && 100.00 & 2.20 & 2.62 && 15.92 & 38.87\\
    \midrule
    \multirow{4}{*}{\textbf{0-3}} & 10\% & 0.84\% & 99.95 & 2.70 & 2.79 && 99.94 & 2.25 & 2.64 && 15.03 & 37.17\\
    & 20\% & 1.69\% & 96.46 & 7.24 & 11.21 && 98.49 & 6.79 & 9.67 && 22.15 & 45.61\\
    & 30\% & 2.53\% & 76.73 & 18.80 & 34.23 && 87.25 & 19.15 & 33.25 && 35.48 & 59.45\\
    & 40\% & 3.38\% & 42.05 & 44.25 & 70.35 && 58.70 & 49.12 & 71.04 && 61.16 & 78.94 \\
    \midrule
    \multirow{4}{*}{\textbf{4-7}} & 10\% & 0.84\% & 100.00 & 1.81 & 2.19 && 100.00 & 1.27 & 2.02 && 13.88 & 36.51\\
    & 20\% & 1.69\% & 100.00 & 1.32 & 3.04 && 99.99 & 1.28 & 3.26 && 15.07 & 39.14\\
    & 30\% & 2.53\% & 99.95 & 1.88 & 6.23 && 99.98 & 1.89 & 7.12 && 17.12 & 43.09\\
    & 40\% & 3.38\% & 99.95 & 2.77 & 11.71 && 99.90 & 2.86 & 12.81 && 18.40 & 45.86\\
    \midrule
    \multirow{4}{*}{\textbf{8-11}} & 10\% & 0.84\% & 100.00 & 2.11 & 3.94 && 100.00 & 1.63 & 3.38 && 16.10 & 41.42\\
    & 20\% & 1.69\% & 99.95 & 5.73 & 22.47 && 99.95 & 5.94 & 23.21 && 22.46 & 55.21\\
    & 30\% & 2.53\% & 99.75 & 21.57 & 62.53 && 99.64 & 21.28 & 62.82 && 35.69 & 75.07\\
    & 40\% & 3.38\% & 98.90 & 33.97 & 72.85 && 99.01 & 33.44 & 72.85 && 45.04 & 79.78\\
    \bottomrule
    \end{tabular}
    }
    %\vspace{-0.2cm}
\end{table}

\subsubsection{Feed-forward Module Pruning}

We perform pruning on all feed-forward modules in the model $M$, as detailed in Tab.~\ref{tab:prune_ff}. It is evident that as the pruning rate increases, both writer classification accuracy and recognition performance decline similarly for the forget and retain sets. This suggests that user-specific and recognition-related information are similarly represented within the feed-forward modules for both the forget and retain sets, indicating a strong coupling between these features. Thus, we choose a pruning rate of 20\% for the feed-forward module to avoid excessive pruning.

\begin{table}[t!]
    \caption{Experiments with varying pruning percentages applied to the feed-forward module.}
    %\vspace{-0.1cm}
    \label{tab:prune_ff}
    \centering
    %\ 
    \small
    \scalebox{0.8}{
    \begin{tabular}{cccccccccccc}
    \toprule
    \multirow{2}{*}{\textbf{Pruning Rate}} & \multirow{2}{*}{\textbf{Sparsity}} & \multicolumn{3}{c}{\textbf{Forget Set}} & & \multicolumn{3}{c}{\textbf{Retain Set}} & & \multicolumn{2}{c}{\textbf{Test Set}}\\
    & & \textbf{ACC} & \textbf{CER} & \textbf{WER} & & \textbf{ACC} & \textbf{CER} & \textbf{WER} & & \textbf{CER} & \textbf{WER}\\
    \midrule
    Original & 0\% & 100.00 & 1.80 & 1.89 && 100.00 & 1.29 & 1.75 && 13.23 & 34.90 \\
    \midrule
    10\% & 5.05\% & 99.80 & 2.58 & 2.84 && 99.23 & 1.81 & 2.64 && 15.05 & 38.01\\
    20\% & 10.13\% & 97.51 & 3.98 & 6.53 && 96.65 & 4.09 & 8.55 && 18.25 & 52.64\\
    30\% & 15.21\% & 89.94 & 12.74 & 22.12 && 88.28 & 13.43 & 25.09 && 27.66 & 52.83\\
    40\% & 20.29\% & 74.74 & 27.94 & 45.29 && 75.07 & 28.32 & 47.31 && 41.05 & 65.31\\
    \bottomrule
    \end{tabular}
    }
    %\vspace{-0.2cm}
\end{table}

\subsubsection{Last Projection Layer Pruning}

We perform pruning on the final projection layer of the writer classification head, as shown in Tab.~\ref{tab:prune_w}. The results demonstrate that pruning can effectively reduce writer classification accuracy to 0 for the forget set while maintaining high accuracy for the retain set. Since only the projection layer of the writer classification head is pruned, recognition performance remains unaffected. This suggests that typical machine unlearning via pruning can work for classification tasks, likely due to the pruning of the final projection layer. However, user-specific knowledge remains embedded throughout the entire model $M$, distributed across all neurons.

In Tab.~\ref{tab:prune_r}, we apply pruning to the final projection layer of the recognition head. As pruning rates increase, recognition performance declines similarly for both the forget and retain sets, while writer classification accuracy remains unaffected, since only the projection layer of the recognition head is pruned.

To avoid relying solely on pruning the final projection layer of the writer classification head, as user-specific knowledge is still retained throughout the model $M$, and to prevent recognition performance degradation, we choose not to prune the final projection layers of either the writer classification or recognition heads.

\subsubsection{Neural Pruning Analysis}

\begin{figure}[ht]
  \centering
  \includegraphics[width=0.95\linewidth]{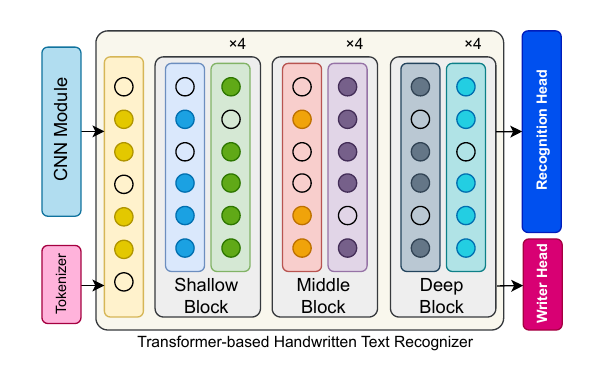}
  \caption{\textbf{Stage I:} Prune 40\% of the embedding module (yellow); 20\%, 40\%, and 20\% of the shallow, middle, and deep self-attention blocks (blue, red, gray), respectively; and 20\% of all feed-forward modules (green, purple, cyan).}
  \label{fig:prune}
\end{figure}

The Neural Pruning stage is illustrated in Fig.~\ref{fig:prune}. After completing all pruning experiments across each module, we can address research question \textbf{[RQ2]} by concluding that neural pruning can partially remove user-identifiable information. However, due to the deep coupling of knowledge within the model for both the forget set and the retain set, encompassing both user-identifiable and recognition-related information, pruning alone is insufficient to fully eliminate user-identifiable information while retaining the model's useful knowledge. Consequently, an additional step is necessary to perform direct unlearning.

While our empirical determination of pruning rates and configurations aimed to systematically uncover patterns related to layer sensitivity and user-identifiable information, it was not intended to suggest repetitive manual tuning in future scenarios. Instead, our comprehensive analysis serves as a foundation for developing more principled, automated pruning strategies. The documented patterns from our study can guide future research, significantly streamlining the unlearning process in new contexts. Additionally, we recognize that neural pruning alone does not fully separate information between the forget and retain sets, as indicated by the concurrent increase in recognition errors. This deep coupling highlights the need for integrating neural pruning with subsequent stages, such as random labeling, to effectively remove user-specific information while preserving recognition accuracy, thereby underscoring their complementary roles in achieving desired privacy outcomes.

\begin{table}[t!]
    \caption{Experiments with varying pruning percentages applied to the final projection layer of the writer classification head.}
    %\vspace{-0.1cm}
    \label{tab:prune_w}
    \centering
    %\ 
    \small
    \scalebox{0.8}{
    \begin{tabular}{cccccccccccc}
    \toprule
    \multirow{2}{*}{\textbf{Pruning Rate}} & \multirow{2}{*}{\textbf{Sparsity}} & \multicolumn{3}{c}{\textbf{Forget Set}} & & \multicolumn{3}{c}{\textbf{Retain Set}} & & \multicolumn{2}{c}{\textbf{Test Set}}\\
    & & \textbf{ACC} & \textbf{CER} & \textbf{WER} & & \textbf{ACC} & \textbf{CER} & \textbf{WER} & & \textbf{CER} & \textbf{WER}\\
    \midrule
    Original & 0\% & 100.00 & 1.80 & 1.89 && 100.00 & 1.29 & 1.75 && 13.23 & 34.90 \\
    \midrule
    10\% & 0.02\% & 0.00 & 1.80 & 1.89 && 100.00 & 1.29 & 1.75 && 13.23 & 34.90\\
    20\% & 0.04\% & 0.00 & 1.80 & 1.89 && 95.25 & 1.29 & 1.75 && 13.23 & 34.90\\
    30\% & 0.06\% & 0.00 & 1.80 & 1.89 && 90.80 & 1.29 & 1.75 && 13.23 & 34.90\\
    40\% & 0.08\% & 0.00 & 1.80 & 1.89 && 85.90 & 1.29 & 1.75 && 13.23 & 34.90\\
    \bottomrule
    \end{tabular}
    }
    %\vspace{-0.2cm}
\end{table}

\begin{table}[t!]
    \caption{Experiments with varying pruning percentages applied to the final projection layer of the recognition head.}
    %\vspace{-0.1cm}
    \label{tab:prune_r}
    \centering
    %\ 
    \small
    \scalebox{0.8}{
    \begin{tabular}{cccccccccccc}
    \toprule
    \multirow{2}{*}{\textbf{Pruning Rate}} & \multirow{2}{*}{\textbf{Sparsity}} & \multicolumn{3}{c}{\textbf{Forget Set}} & & \multicolumn{3}{c}{\textbf{Retain Set}} & & \multicolumn{2}{c}{\textbf{Test Set}}\\
    & & \textbf{ACC} & \textbf{CER} & \textbf{WER} & & \textbf{ACC} & \textbf{CER} & \textbf{WER} & & \textbf{CER} & \textbf{WER}\\
    \midrule
    Original & 0\% & 100.00 & 1.80 & 1.89 && 100.00 & 1.29 & 1.75 && 13.23 & 34.90 \\
    \midrule
    5\% & 0.00\% & 100.00 & 1.90 & 2.34 && 100.00 & 1.32 & 1.92 && 13.22 & 34.88\\
    10\% & 0.00\% & 100.00 & 21.26 & 57.60 && 100.00 & 18.78 & 54.75 && 28.26 & 67.62\\
    15\% & 0.01\% & 100.00 & 22.82 & 61.53 && 100.00 & 20.20 & 57.90 && 29.26 & 69.06\\
    20\% & 0.01\% & 100.00 & 26.58 & 64.97 && 100.00 & 23.65 & 61.48 && 32.20 & 71.00\\
    25\% & 0.01\% & 100.00 & 36.79 & 82.11 && 100.00 & 33.48 & 78.46 && 40.69 & 83.66\\
    30\% & 0.01\% & 100.00 & 526.64 & 100.00 && 100.00 & 521.10 & 100.00 && 539.53 & 100.00\\
    %40\% & 0.01\% & 100.00 & 532.23 & 100.00 && 100.00 & 527.00 & 100.00 && 545.16 & 100.00\\
    \bottomrule
    \end{tabular}
    }
    %\vspace{-0.2cm}
\end{table}

\subsection{Final Results using the prune-unlearn pipeline with sota MU methods}

In this section, we evaluate the effectiveness of our prune-unlearn pipeline when integrated with several state-of-the-art machine unlearning techniques: Random Labeling, Fisher Forgetting, Amnesiac Unlearning, DELETE, and our proposed WIC method. We compare these approaches based on their ability to remove user-identifiable information while maintaining handwriting text recognition performance. To assess the level of unlearning, we perform membership inference attacks (MIA) on all methods. The goal is for the forget set to behave similarly to the test set, indicating non-membership, while the retain set continues to exhibit characteristics of training members.

For Random Labeling, we conduct more detailed experiments across different unlearning iterations. To ensure a fair and consistent comparison across all methods, we fix the total number of training iterations to 10,000 and use a batch size of 64 throughout.

\subsubsection{Random Labeling Experiments}

Both the baseline model $M$ and the pruned model $M^*$ are then fine-tuned on the updated training set $D_{\text{train}}'$, which comprises the randomly labeled forget set $D_{\text{forget}}'$, where the user IDs in the forget set are replaced with random user IDs, excluding the real ones, and the retain set $D_{\text{retain}}$. The results are shown for different iterations as detailed in Tab.~\ref{tab:rl}.

The results indicate that, following pruning, the pruned model outperforms the baseline model at the same iterations of random labeling. At epoch 10,000, the pruned model achieves 0\% writer classification accuracy, signifying a complete forgetting of user-specific information in the forget set, while retaining a high accuracy of 99.95\% for the retain set. In contrast, the baseline model requires 38,000 iterations to reach 0\% writer classification accuracy. Regarding recognition performance, the prune-first-then-random-label method achieves superior results for the forget set, with a CER of 0.28\% and a WER of 1.05\%, compared to the baseline method, which requires more iterations and results in a CER of 3.73\% and a WER of 2.59\%.

\begin{table}[ht]
    \caption{Experiments with random labeling applied to both the baseline and pruned models. }
    %\vspace{-0.1cm}
    \label{tab:rl}
    \centering
    %\ 
    \small
    \scalebox{0.8}{
    \begin{tabular}{ccccccccccccc}
    \toprule
    \multirow{2}{*}{\textbf{Method}} & \multirow{2}{*}{\textbf{Iter.}} & \multirow{2}{*}{\textbf{Sparsity}} & \multicolumn{3}{c}{\textbf{Forget Set}} & & \multicolumn{3}{c}{\textbf{Retain Set}} & & \multicolumn{2}{c}{\textbf{Test Set}}\\
    & & & \textbf{ACC} & \textbf{CER} & \textbf{WER} & & \textbf{ACC} & \textbf{CER} & \textbf{WER} & & \textbf{CER} & \textbf{WER}\\
    \midrule
    Baseline $M$ & 0 & 0\% & 100.00 & 1.80 & 1.89 && 100.00 & 1.29 & 1.75 && 13.23 & 34.90 \\
    +RL & 1,000 & 0\% & 7.13 & 2.76 & 3.49 && 99.96 & 3.16 & 3.64 && 15.56 & 37.66\\
    +RL & 5,000 & 0\% & 1.35 & 2.69 & 3.54 && 99.97 & 4.39 & 4.66 && 17.43 & 39.69\\
    +RL & 10,000 & 0\% & 0.15 & 3.99 & 4.33 && 99.99 & 4.32 & 4.36 && 16.99 & 38.60\\
    +RL & 38,000 & 0\% & 0.00 & 3.73 & 2.59 && 100.00 & 2.73 & 2.95 && 14.87 & 36.64\\
    \midrule
    %0.1-0.1-0.3-0.1-0.1 & 9.40\% & 92.28 & 9.82 & 30.54 && 95.86 & 10.73 & 31.98 && 26.52 & 56.69\\
    Pruned $M^*$ & 0 & 17.45\% & 20.03 & 58.29 & 84.16 && 35.50 & 60.44 & 85.26 && 66.81 & 88.32\\
    +RL & 1,000 & 7.49\% & 3.49 & 0.73 & 1.40 && 99.08 & 2.43 & 9.41 && 15.66 & 40.67\\
    +RL & 5,000 & 7.44\% & 0.75 & 0.67 & 1.10 && 99.84 & 2.44 & 3.97 && 15.71 & 38.46\\
    +RL & 10,000 & 7.41\% & 0.00 & 0.28 & 1.05 && 99.95 & 2.46 & 3.27 && 14.98 & 37.61\\
    \bottomrule
    \end{tabular}
    }
    %\vspace{-0.2cm}
\end{table}

\textbf{Membership Inference Evaluation}

As previously discussed, the writer classification head serves as an indicator. However, achieving 0\% accuracy does not necessarily confirm non-membership for the forget set. To address this, we conduct a membership inference evaluation, as shown in Tab.~\ref{tab:mia}. The forget set is initially part of the member category in the baseline recognition model, since it is included in the training data. After applying unlearning, our goal is for the forget set to exhibit behavior similar to non-members, effectively transitioning to an unseen status in the unlearned model. For both the baseline and pruned models, with and without random labeling, members (retain set) are classified as ``seen'' with over 75\% accuracy, while non-members (test set) are classified as ``unseen'' with an accuracy range of 51\% to 59\%. These outcomes align with expectations: members have a high classification probability as they were seen during training, while non-members approach the random guessing baseline (50\%), reflecting the model's lack of prior knowledge about them.

\begin{table}[ht]
    \caption{Membership inference evaluation with random labeling applied to both the baseline and pruned models. Forget Set is member in the baseline; pruning and RL aim to make it appear as non-member.}
    %\vspace{-0.1cm}
    \label{tab:mia}
    \centering
    %\ 
    \small
    \scalebox{0.8}{
    \begin{tabular}{ccccccccccc}
    \toprule
    \multirow{2}{*}{\textbf{Method}} & \multirow{2}{*}{\textbf{Iter.}} & \multicolumn{2}{c}{\textbf{Forget Set}} & & \multicolumn{2}{c}{\textbf{Members (Retain)}} & & \multicolumn{2}{c}{\textbf{Non-members (Test)}}\\
    & & \textbf{Seen} & \textbf{Unseen} && \textbf{Seen} & \textbf{Unseen} && \textbf{Seen} & \textbf{Unseen}\\
    \midrule
    Baseline $M$ & 0 & 73.29 & 26.71 && 80.92 & 19.08 && 45.61 & 54.39\\
    +RL & 1,000 & 67.16 & 32.84 && 83.60 & 16.40 && 45.24 & 54.76\\
    +RL & 5,000 & 56.85 & 43.15 && 82.90 & 17.10 && 45.70 & 54.30\\
    +RL & 10,000 & 51.47 & 48.53 && 82.63 & 17.37 && 45.67 & 54.33\\
    +RL & 38,000 & 46.29 & 53.71 && 84.66 & 15.34 && 44.06 & 55.94\\
    \midrule
    Pruned $M^*$ & 0 & 64.42 & 35.58 && 75.85 & 24.15 && 48.24 & 51.76\\
    +RL & 1,000 & 61.63 & 38.37 && 75.76 & 24.24 && 43.45 & 56.55\\
    +RL & 5,000 & 53.86 & 46.14 && 76.17 & 23.83 && 41.60 & 58.40\\
    +RL & 10,000 & 49.98 & 50.02 && 77.74 & 22.26 && 41.58 & 58.42\\
    \bottomrule
    \end{tabular}
    }
    %\vspace{-0.2cm}
\end{table}

For the forget set, the baseline model classifies it as ``seen'' members with a probability of 73.29\%, since the forget set is part of the training data for model $M$. In contrast, the pruned model $M^*$ reduces this probability to 64.42\%, indicating that pruning effectively removes some user-specific information. Comparing random labeling results for the baseline and pruned models at the same iterations, the pruned model retains less user-specific information, with its results closer to the random guess threshold of 50\%. At iteration 10,000, random labeling on the pruned model achieves a classification probability of 49.98\% for ``seen'' members, confirming that user-specific information has been effectively removed from model $M'$.

Thus, we can now address research question \textbf{[RQ3]} by concluding that using machine unlearning techniques in the writer classification head effectively removes user-identifiable information without negatively impacting recognition performance.

\subsubsection{Fisher Forgetting Experiments}

Fisher Forgetting is applied once over the entire forget set to compute both the gradient $g_f$ and the diagonal Fisher matrix. Since the forget set contains 2,007 samples and the batch size is 64, this step is equivalent to approximately 32 iterations. Given the total budget of 10,000 iterations, we use 32 iterations for Fisher Forgetting and allocate the remaining 9,968 iterations for fine-tuning.

\begin{table}[t!]
    \caption{Experiments with Fisher Forgetting applied to both the baseline and pruned models. }
    %\vspace{-0.1cm}
    \label{tab:fisher}
    \centering
    \small
    \scalebox{0.8}{
    \begin{tabular}{ccccccccccccc}
    \toprule
    \multirow{2}{*}{\textbf{Method}} & \multirow{2}{*}{\textbf{Iter.}} & \multirow{2}{*}{\textbf{Sparsity}} & \multicolumn{3}{c}{\textbf{Forget Set}} & & \multicolumn{3}{c}{\textbf{Retain Set}} & & \multicolumn{2}{c}{\textbf{Test Set}}\\
    & & & \textbf{ACC} & \textbf{CER} & \textbf{WER} & & \textbf{ACC} & \textbf{CER} & \textbf{WER} & & \textbf{CER} & \textbf{WER}\\
    \midrule
    Baseline $M$ & 0 & 0\% & 100.00 & 1.80 & 1.89 && 100.00 & 1.29 & 1.75 && 13.23 & 34.90 \\
    +Fisher & 32 & 0\% & 5.38 & 52.45 & 76.63 && 6.02 & 50.60 & 75.95 && 60.63 & 81.41 \\
    +FT & 9,968 & 0\% & 7.67 & 11.73 & 8.72 && 99.98 & 8.64 & 8.40 && 25.08 & 47.65 \\
    \midrule
    Pruned $M^*$ & 0 & 17.45\% & 20.03 & 58.29 & 84.16 && 35.50 & 60.44 & 85.26 && 66.81 & 88.32\\
    +Fisher & 32 & 8.28\% &  6.08 & 584.07 & 100.00 && 6.37 &  581.22 & 100.00 && 593.86 & 100.00\\
    +FT & 9,968 &  7.39\% & 8.87 & 5.21 & 9.77 && 99.99 & 2.52 & 3.82 && 15.16 & 37.38 \\
    \bottomrule
    \end{tabular}
    }
    %\vspace{-0.2cm}
\end{table}

From Table~\ref{tab:fisher}, we observe that applying Fisher Forgetting alone leads to a performance drop in accuracy, CER, and WER for both the forget set and the retain set. This degradation is likely due to the highly entangled nature of information in sequence recognition models, which makes targeted forgetting more challenging compared to simpler classification tasks.

To recover overall performance after unlearning, the model is further fine-tuned on the retain set. Surprisingly, despite the forget set being excluded during this fine-tuning phase, the writer classification accuracy on the forget set slightly improves. This indicates that Fisher Forgetting alone may not have fully removed the forget set's influence. One possible explanation is that latent representations associated with the forget set remain in the model and are unintentionally reinforced during retain-set fine-tuning. 

Moreover, the MIA success rate on the forget set is 64\%, which is above the random guessing baseline of 50\%. This indicates a failure to fully achieve the unlearning objective from a privacy perspective, as the model still partially treats the forget set as training members. While Fisher Forgetting may reduce some direct memorization, residual user-identifiable signals persist, highlighting its limitations in fully erasing sensitive information.

\subsubsection{Amnesiac Unlearning Experiments}

Amnesiac Unlearning~\cite{graves2021amnesiac} reverses the influence of the forget set by applying negative gradient updates scaled by a factor \(\alpha\), effectively approximating the inverse of the original training dynamics. The scaling factor \(\alpha\) is set to \(1 \times 10^{-6}\).

% \begin{equation}
% \mathcal{L}_{\text{amnesiac}} = -\alpha \cdot \mathcal{L}_{\text{wid}} 
% \end{equation}

% Here, \(\mathcal{L}_{\text{wid}}\) denotes the writer classification loss. Note that we exclude the recognition loss in this formulation.

\begin{table}[ht]
    \caption{Experiments with Amnesiac Unlearning applied to both the baseline and pruned models. }
    %\vspace{-0.1cm}
    \label{tab:amnesiac}
    \centering
    \small
    \scalebox{0.8}{
    \begin{tabular}{ccccccccccccc}
    \toprule
    \multirow{2}{*}{\textbf{Method}} & \multirow{2}{*}{\textbf{Iter.}} & \multirow{2}{*}{\textbf{Sparsity}} & \multicolumn{3}{c}{\textbf{Forget Set}} & & \multicolumn{3}{c}{\textbf{Retain Set}} & & \multicolumn{2}{c}{\textbf{Test Set}}\\
    & & & \textbf{ACC} & \textbf{CER} & \textbf{WER} & & \textbf{ACC} & \textbf{CER} & \textbf{WER} & & \textbf{CER} & \textbf{WER}\\
    \midrule
    Baseline $M$ & 0 & 0\% & 100.00 & 1.80 & 1.89 && 100.00 & 1.29 & 1.75 && 13.23 & 34.90 \\
    +Amnesiac & 32 & 0\% & 3.14 & 69.28 & 89.24 && 0.10 & 68.35 & 88.52 && 76.05 & 91.60\\
    +FT & 9,968 & 0\% &  4.79 & 7.48 & 7.42 && 99.96 & 4.92 & 5.56 && 20.45 & 45.18\\
    \midrule
    Pruned $M^*$ & 0 & 17.45\% & 20.03 & 58.29 & 84.16 && 35.50 & 60.44 & 85.26 && 66.81 & 88.32\\
    +Amnesiac & 32 & 7.84\% &  0.00 & 88.38 & 98.75 && 0.39 & 89.18 & 98.71 && 89.44 & 98.57\\
    +FT & 9,968 &  7.40\% & 7.47 & 4.88 & 8.87 && 99.99 & 3.92 & 4.87 && 16.53 & 38.77\\
    \bottomrule
    \end{tabular}
    }
    %\vspace{-0.2cm}
\end{table}

From Table~\ref{tab:amnesiac}, we observe that applying Amnesiac Unlearning alone leads to a drop in accuracy, CER, and WER for both the forget set and the retain set. This performance degradation highlights the highly entangled nature of learned representations between the forget and retain sets in handwriting recognition tasks. The difficulty of isolating user-specific information without affecting shared representations makes effective unlearning particularly challenging in this context.

Interestingly, we find that the pruning stage significantly strengthens the impact of Amnesiac Unlearning. Specifically, pruning leads to a strong removal of information, reducing the writer classification accuracy on the forget set to 0\% at this point. However, after fine-tuning the model on the retain set, the accuracy on the forget set increases to 7.47\%. This recovery suggests that, although pruning initially removes the forget set information, some hidden traces may still remain in the model and get activated again during fine-tuning. This shows that Amnesiac Unlearning, even when combined with pruning, may still struggle to fully prevent the return of user-identifiable patterns, likely due to the shared features between the forget and retain sets.

Supporting this observation, the MIA success rate on the forget set reaches 67\%, which is significantly above the random baseline of 50\%. This result indicates that the model continues to treat the forget set as part of its training data, confirming that the unlearning process is incomplete from a privacy standpoint.

\subsubsection{DELETE Experiments}

We further evaluate the effectiveness of the DELETE method~\cite{zhou2025decoupled} in our prune-unlearn pipeline. To ensure a fair comparison with other machine unlearning techniques, we follow the same experimental setup by limiting the total training iterations to 10,000. Specifically, we allocate 5,000 iterations to the DELETE unlearning phase, during which the model is optimized to remove the influence of the forget set. The remaining 5,000 iterations are dedicated to fine-tuning the model exclusively on the retain set to recover any potential loss in task performance.

\begin{table}[ht]
    \caption{Experiments with DELETE applied to both the baseline and pruned models. }
    %\vspace{-0.1cm}
    \label{tab:delete}
    \centering
    
    \small
    \scalebox{0.8}{
    \begin{tabular}{ccccccccccccc}
    \toprule
    \multirow{2}{*}{\textbf{Method}} & \multirow{2}{*}{\textbf{Iter.}} & \multirow{2}{*}{\textbf{Sparsity}} & \multicolumn{3}{c}{\textbf{Forget Set}} & & \multicolumn{3}{c}{\textbf{Retain Set}} & & \multicolumn{2}{c}{\textbf{Test Set}}\\
    & & & \textbf{ACC} & \textbf{CER} & \textbf{WER} & & \textbf{ACC} & \textbf{CER} & \textbf{WER} & & \textbf{CER} & \textbf{WER}\\
    \midrule
    Baseline $M$ & 0 & 0\% & 100.00 & 1.80 & 1.89 && 100.00 & 1.29 & 1.75 && 13.23 & 34.90 \\
    +DELETE & 5,000 & 0\% & 0.40 & 21.70 & 41.16 && 99.93 & 16.84 & 26.48 && 36.05 & 65.03\\
    +FT & 5,000 & 0\% & 0.00 & 7.54 & 7.67 && 99.99 & 7.16 & 7.25 && 22.11 & 46.06\\
    \midrule
    Pruned $M^*$ & 0 & 17.45\% & 20.03 & 58.29 & 84.16 && 35.50 & 60.44 & 85.26 && 66.81 & 88.32\\
    +DELETE & 5,000 & 7.48\%  & 0.00 & 62.26 & 87.54 && 92.85 & 61.57 & 87.82 && 70.24 & 90.89\\
    +FT & 5,000 &  7.43\% & 0.00 & 7.01 & 11.16 && 99.99 & 4.70 & 5.65 && 17.74 & 39.95\\
    \bottomrule
    \end{tabular}
    }
    %\vspace{-0.2cm}
\end{table}

From Table~\ref{tab:delete}, we observe that DELETE is effective in unlearning writer ID information, achieving only 0.40\% classification accuracy on the forget set when applied to the base model $M$, and dropping further to 0.00\% when applied to the pruned model $M^*$. At the same time, it maintains high accuracy on the retain set, demonstrating strong task retention. Although both CER and WER initially degrade for both the forget and retain sets, the performance significantly improves after the fine-tuning stage, showing the potential of DELETE in handwriting recognition unlearning scenarios. Furthermore, the MIA success rate on the forget set is 46\%, which is close to the random baseline of 50\%, indicating successful unlearning from a privacy perspective.

\subsubsection{Writer-ID Confusion (WIC) Experiments}

Finally, we conduct experiments using our proposed Writer-ID Confusion (WIC) method. To determine the optimal value for the confusion loss weight $\lambda$, we perform a hyperparameter search over 2,000 training iterations, as shown in Table~\ref{tab:wic}. The results show that varying $\lambda$ from 0.2 to 0.8 yields generally stable performance across the forget, retain, and test sets. Among the results, $\lambda = 0.4$ achieves the best balance between effective forgetting and accuracy preservation, and is therefore used in all subsequent WIC experiments.

\begin{table}[ht]
    \caption{Hyper-parameter search for our method WIC on pruned models. }
    %\vspace{-0.1cm}
    \label{tab:wic}
    \centering
    \small
    \scalebox{0.8}{
    \begin{tabular}{ccccccccccccc}
    \toprule
    \multirow{2}{*}{\textbf{$\lambda$}} & \multirow{2}{*}{\textbf{Iter.}} & \multirow{2}{*}{\textbf{Sparsity}} & \multicolumn{3}{c}{\textbf{Forget Set}} & & \multicolumn{3}{c}{\textbf{Retain Set}} & & \multicolumn{2}{c}{\textbf{Test Set}}\\
    & & & \textbf{ACC} & \textbf{CER} & \textbf{WER} & & \textbf{ACC} & \textbf{CER} & \textbf{WER} & & \textbf{CER} & \textbf{WER}\\
    \midrule
    0.2 & 2,000 & 7.47\% & 0.10 & 1.12 & 1.54 & & \textbf{99.82} & 2.19 & 5.71 & & 15.37 & \textbf{39.63} \\
    0.4 & 2,000 & 7.47\% & \textbf{0.00} & 0.81 & \textbf{1.00} & & \textbf{99.82} & \textbf{1.79} & \textbf{5.09} & & 15.57 & 39.80\\
    0.6 & 2,000 & 7.46\% & \textbf{0.00} & 0.59 & 1.35 & & 99.69 & 1.90 & 6.09 & & 15.81 & 40.83 \\
    0.8 & 2,000 & 7.46\% & \textbf{0.00} & \textbf{0.42} & 1.30 & & 99.54 & 1.80 & 6.25 & & \textbf{15.35} & 40.13\\
    \bottomrule
    \end{tabular}
    }
    %\vspace{-0.2cm}
\end{table}

\subsubsection{Final Comparison with State-of-the-Art Methods}

Since our proposed WIC method does not require an additional fine-tuning stage, we directly report the prune-WIC results in Table~\ref{tab:final}. As shown, even with just 2,000 iterations, WIC achieves state-of-the-art performance. When extended to 10,000 iterations, it delivers results that are comparable to or even better than existing methods. To assess the statistical significance of the observed performance differences, we perform a Multivariate Analysis of Variance (MANOVA) using the joint metrics of ACC, CER, and WER, and confirm that the improvements are statistically significant. Furthermore, the MIA success rates on the forget set are 49\% and 48\% after 2,000 and 10,000 iterations, respectively. Both are close to the random guess baseline of 50\%, indicating effective unlearning from a privacy standpoint.

\begin{table}[ht]
    \caption{Final comparison with state-of-the-art methods on the IAM datasets.}
    %\vspace{-0.1cm}
    \label{tab:final}
    \centering
  
    \small
    \scalebox{0.8}{
    \begin{tabular}{ccccccccccccc}
    \toprule
    \multirow{2}{*}{\textbf{Method}} & \multirow{2}{*}{\textbf{Iter.}} & \multirow{2}{*}{\textbf{Sparsity}} & \multicolumn{3}{c}{\textbf{Forget Set}} & & \multicolumn{3}{c}{\textbf{Retain Set}} & & \multicolumn{2}{c}{\textbf{Test Set}}\\
    & & & \textbf{ACC} & \textbf{CER} & \textbf{WER} & & \textbf{ACC} & \textbf{CER} & \textbf{WER} & & \textbf{CER} & \textbf{WER}\\
    \midrule
    Baseline $M$ & 0 & 0\% & 100.00 & 1.80 & 1.89 && 100.00 & 1.29 & 1.75 && 13.23 & 34.90 \\
    Pruned $M^*$ & 0 & 17.45\% & 20.03 & 58.29 & 84.16 && 35.50 & 60.44 & 85.26 && 66.81 & 88.32\\
    \midrule
    Prune-RL & 10,000 & 7.41\% & \textbf{0.00} & \textbf{0.28} & 1.05 && 99.95 & 2.46 & 3.27 && 14.98 & 37.61\\
    Prune-Fisher-FT & 10,000 &  7.39\% & 8.87 & 5.21 & 9.77 && \textbf{99.99} & 2.52 & 3.82 && 15.16 &  38.14\\
    Prune-Amnesiac-FT & 10,000 &  7.40\% & 7.47 & 4.88 & 8.87 && \textbf{99.99} & 3.92 & 4.87 && 16.53 & 38.77\\
    Prune-DELETE-FT & 10,000 &  7.43\% & \textbf{0.00} & 7.01 & 11.16 && \textbf{99.99} & 4.70 & 5.65 && 17.74 & 39.95\\
    \midrule
    \textbf{Prune-WIC (Ours)} & 2,000 & 7.47\% & \textbf{0.00} & 0.81 & \textbf{1.00} && 99.82 & 1.79 & 5.09 && 15.57 & 39.80\\
    \textbf{Prune-WIC (Ours)} & 10,000 & 7.42\% & \textbf{0.00} & 0.97 & 1.35 && 99.96 & \textbf{1.75} & \textbf{3.05} && \textbf{14.83} & \textbf{37.38}\\
    \bottomrule
    \end{tabular}
    }
    %\vspace{-0.2cm}
\end{table}

\subsubsection{Final Comparison on the CVL datasets}

The CVL datasets comprise a training set of 27 writers and a test set of 283 writers. To evaluate the robustness of our proposed method, we merge all the writers, resulting in a total of 310 writers and 99,904 word-level images. This combined dataset is used for unlearning comparison. We limit the unlearning process to a maximum of 4,000 iterations. The final performance comparison is presented in Table~\ref{tab:cvl}.

Our proposed method, Prune-WIC, achieves a compelling balance between effective forgetting and performance retention. On the Forget Set, Prune-WIC attains the lowest writer identification accuracy (ACC) of 0.00 and a WER of 4.68\%, outperforming all baseline methods. In particular, it shows significant improvement compared to Prune-Fisher-FT and Prune-DELETE-FT, which suffer from higher CER and WER values. On the Retain Set, Prune-WIC maintains strong retention performance with an accuracy of 98.36\%, a CER of 5.70\%, and a WER of 7.16\%, exhibiting only marginal degradation compared to the original Baseline model. 

To assess the statistical significance of the observed performance differences, we perform a Multivariate Analysis of Variance (MANOVA) using the joint metrics of ACC, CER, and WER, and confirm that the improvements are statistically significant.

In terms of computational efficiency, Prune-WIC completes the unlearning process in 4,069 seconds while using 16.4 GB of GPU memory, demonstrating a favorable trade-off between effectiveness and resource consumption. These results confirm the effectiveness and practicality of our approach for unlearning in handwriting recognition scenarios.

\begin{table}[ht]
    \caption{Final comparison with state-of-the-art methods on the CVL datasets.}
    %\vspace{-0.1cm}
    \label{tab:cvl}
    \centering
    \ 
    \small
    \scalebox{0.75}{
    \begin{tabular}{ccccccccccccc}
    \toprule
    \multirow{2}{*}{\textbf{Method}} & \multirow{2}{*}{\textbf{Iter.}} & \multirow{2}{*}{\textbf{Sparsity}} & \multicolumn{3}{c}{\textbf{Forget Set}} & & \multicolumn{3}{c}{\textbf{Retain Set}} && \textbf{Time} &  \textbf{Peak Mem.} \\
    & & & \textbf{ACC} & \textbf{CER} & \textbf{WER} & & \textbf{ACC} & \textbf{CER} & \textbf{WER} && \textbf{(second)} & \textbf{GPU (GB)}\\
    \midrule
    Baseline $M$ & 0 & 0\% & 100.00 & 4.91 & 5.22 && 100.00 & 6.47 & 5.70 && $-$ & $-$\\
    Pruned $M^*$ & 0 & 17.44\% & 5.29 & 65.40 & 86.47 && 14.33 & 68.70 & 88.00 && 3432 & 13.8\\
    \midrule
    Prune-RL & 4,000 & 7.40\% & 0.25 & 3.81 & 6.55 && 92.04 & 6.93 & 10.35 && 6362 & \textbf{16.4}\\
    Prune-Fisher-FT & 4,000 &  7.30\% & \textbf{0.00} & 42.72 & 72.66 && 33.17 & 42.26 & 72.17 && \textbf{1025} & \textbf{16.4}\\
    Prune-Amnesiac-FT & 4,000 & 7.40\% & 9.17 & \textbf{3.68} & 8.00 && 96.49 & \textbf{4.21} & 7.36 && 1029 & \textbf{16.4}\\
    Prune-DELETE-FT & 4,000 & 7.40\% & 0.00 & 7.34 & 10.62 && 97.47 & 8.21& 9.24 && 1740 & 17.8\\
    \midrule
    \textbf{Prune-WIC (Ours)} & 4,000 & 7.39\% & \textbf{0.00} & 4.68 & \textbf{6.08} && \textbf{98.36} & 5.70 & \textbf{7.16} && 4069 & \textbf{16.4}\\
    \bottomrule
    \end{tabular}
    }
    %\vspace{-0.2cm}
\end{table}

\section{Conclusion}

{\ {
In this work, we propose a novel two-stage machine unlearning framework for transformer-based handwriting text recognition (HTR) systems, focusing on removing user-identifiable information while preserving model utility. The first stage introduces a targeted neural pruning strategy that analyzes neuron importance based on activation ratios between the forget and retain sets. This method not only enables effective removal of sensitive information but also provides practical insights into where and how to prune the model. Importantly, this pruning strategy can serve as a general guideline for future unlearning tasks, eliminating the need to repeatedly re-run expensive experiments to determine which blocks to prune or at what probability.

In the second stage, we introduce Writer-ID Confusion (WIC), our key contribution to machine unlearning. WIC enforces a uniform distribution over writer predictions for the forget set while retaining standard training on the retain set. This lightweight, plug-and-play method achieves competitive or superior performance compared to state-of-the-art unlearning techniques, all without requiring additional fine-tuning or excessive computation. WIC proves to be a practical and effective solution for real-world privacy-preserving HTR applications.

While our work primarily focuses on writer identity, it does not yet address other sources of information leakage such as rare lexical patterns or unique stylistic traits. Future research should explore these aspects using techniques like stylometry or lexical distribution analysis to broaden privacy coverage.

Additionally, this framework opens several promising research directions. Future work could explore layout-aware machine unlearning for selectively forgetting content in documents with complex spatial structures, or multilingual unlearning to address language-specific forgetting and cross-linguistic effects. Applying these approaches to downstream tasks such as document visual question answering (DocVQA)~\cite{tito2021icdar} could enhance privacy by enabling systems to forget entire documents from retrievable memory. Integrating the proposed prune-unlearn mechanisms into retrieval-based DocVQA systems~\cite{kang2024multi} would allow models to be updated upon user requests, preventing access to specific documents and ensuring stronger end-to-end privacy guarantees.

}

\section*{Acknowledgements}

This work has been supported by the Beatriu de Pinós del Departament de Recerca i Universitats de la Generalitat de Catalunya (2022 BP 00256), the predoctoral program AGAUR-FI ajuts (2024 FI-3 00065) Joan Oró, which is backed by the Secretariat of Universities and Research of the Department of Research and Universities of the Generalitat of Catalonia, as well as the European Social Plus Fund, European Lighthouse on Safe and Secure AI (ELSA) from the European Union’s Horizon Europe programme under grant agreement No 101070617, Ramon y Cajal research fellowship RYC2020-030777-I / AEI / 10.13039/501100011033. This work has also been supported by Riksbankens Jubileumsfond, grant M24-0028 (Echoes of History: Analysis and Decipherment of Historical Writings, DESCRYPT), the Spanish projects CNS2022-135947 (DOLORES), PID2021-126808OB-I00 (GRAIL) and PID2024-157778OB-I00 (SUKIDI), the Consolidated Research Group 2021 SGR 01559 from the Research and University Department of the Catalan Government, and PID2023-146426NB-100 funded by MCIU/AEI/10.13039/501100011033 and FSE+.

\bibliographystyle{elsarticle-num}
\bibliography{bib}
% \begin{thebibliography}{00}

% %% For numbered reference style
% %% \bibitem{label}
% %% Text of bibliographic item

% \bibitem{lamport94}
%   Leslie Lamport,
%   \textit{\LaTeX: a document preparation system},
%   Addison Wesley, Massachusetts,
%   2nd edition,
%   1994.

% \end{thebibliography}
\end{document}